\documentclass[conference]{IEEEtran}

\usepackage{setspace}

\usepackage{numprint}
\npdecimalsign{.}
\nprounddigits{3}

\usepackage{multirow}

\usepackage{lipsum}
\usepackage{algorithm}
\usepackage{algorithmic,eqparbox,array}

\usepackage{balance}

\usepackage{amscd}
\usepackage{mathtools}

\usepackage{amsmath,amssymb,amsfonts}
\usepackage{algorithmic}
\usepackage{graphicx}
\usepackage{textcomp}
\usepackage{xcolor}
\def\BibTeX{{\rm B\kern-.05em{\sc i\kern-.025em b}\kern-.08em
    T\kern-.1667em\lower.7ex\hbox{E}\kern-.125emX}}

\begin{document}

\title{\huge \bf Generalized Image Reconstruction over T-Algebra
\thanks{Identify applicable funding agency here. If none, delete this.}
}

\author{\IEEEauthorblockN{Liang Liao}
\IEEEauthorblockA{{School of Electronics and Information} \\
{Zhongyuan University of Technology}\\
Zhengzhou, China \\
liaoliangis@126.com}
\and
\IEEEauthorblockN{Xuechun Zhang}
\IEEEauthorblockA{{School of Electronics and Information} \\
{Zhongyuan University of Technology}\\
Zhengzhou, China \\
zhangxuechun2020@126.com}
\and
\IEEEauthorblockN{Xinqiang Wang}
\IEEEauthorblockA{{School of Electronics and Information} \\
{Zhongyuan University of Technology}\\
Zhengzhou, China \\
dqxwxq@sina.com}
\and 
\IEEEauthorblockN{~\hspace{0.01em}}
\IEEEauthorblockA{{~\hspace{0.01em}~} \\
}
\and 
\IEEEauthorblockN{~\hspace{-5em}}
\IEEEauthorblockA{{~\hspace{-5em}~} \\
}
\and
\IEEEauthorblockN{Sen Lin}
\IEEEauthorblockA{{School of Electronics and Information} \\
{Zhongyuan University of Technology}\\
Zhengzhou, China \\
lins16@126.com}
\and
\IEEEauthorblockN{Xin Liu}
\IEEEauthorblockA{{Information Center} \\
{Yellow River Conservancy Commission}\\
Zhengzhou, China \\
liuxinis@126.com}
}

\maketitle

\begin{abstract}
Principal Component Analysis (PCA) is well known for its capability of dimension reduction and data compression. However, when using PCA for compressing/reconstructing images,  images need to be recast to vectors. The vectorization of images makes some correlation constraints of neighboring pixels and spatial information lost.  To deal with the drawbacks of the vectorizations adopted by PCA,  we used small neighborhoods of each pixel to form compoun pixels and use a tensorial version of PCA, called TPCA (Tensorial Principal Component Analysis), to compress and reconstruct a compound image of compound pixels. Our experiments on public data show that TPCA compares favorably with PCA in compressing and reconstructing images.  We also show in our experiments that the performance of TPCA increases when the order of compound pixels increases. 
\end{abstract}

\begin{IEEEkeywords}
generalized image analysis, tensorial algebra, principal component analysis, compound image
\end{IEEEkeywords}


\section{Introduction}
Principal component analysis (PCA), well-known for dimension reduction and information extraction, is widely used in machine intelligence, data compression, image analysis, etc.

Due to the wide-spread of PCA, many authors try to improve PCA. A PCA variant called 2DPCA (Two-Dimensional PCA) is proposed by Yang et al. with face recognition applications.
The covariance matrix size for computing the eigenvectors in 2DPCA is much smaller than that of PCA \cite{yang2004two}.
Zhang et al. propose the (2D)\textsuperscript{2}PCA (Two-Directional Two-Dimensional PCA), which extracts information along the row direction and column direction at the same time, and its recognition rate is claimed to be higher than that of 2DPCA \cite{zhang20052d}. 
Mat-PCA is proposed by Chen et al. Compared with PCA, Mat-PCA computes the covariance matrix of images and calculates the feature vectors more effectively \cite{chen2005feature}.

Although the above methods either reduce the computational complexity or improve computational performance, they do not particularly consider structural constraints at the pixel level.
Whether one can adopt a straightforward and systematical approach to exploiting the spatial constraints of pixels? If the answer is yes, one might need a non-naïve algebraic approach. One genesis of such an approach can be the “t-product” model by M. E. Kilmer et al., which is used to generalize the SVD (Singular Value Decomposition) of a matrix to its third-order version, called t-SVD \cite{kilmer2011factorization,kilmer2013third}. 
Inspired by the ``t-product'' model, Liao and Maybank propose a novel semisimple communicative algebra, called t-algebra, via the multi-way Fourier transform \cite{liao2020generalized,liao2020general}.  
The elements of the t-algebra are represented by fixed-sized high-order complex arrays and generalize complex numbers, therefore, referred to as ``t-scalars.''
With the paradigm shift from over the field of complex numbers to over the recently-proposed t-algebra, the classical principal component analysis (PCA) can be straightforwardly generalized to its higher-order version, called TPCA 
(Tensorial PCA) \cite{liao2020generalized,liao2020general,liao2017hyperspectral}. To demonstrate the t-matrix paradigm and the performance of TPCA, we use TPCA to compress and reconstruct image data.

As a following research of the work in \cite{liao2020generalized,liao2020general}, besides demonstrating the t-matrix paradigm and TPCA’s performance, we also investigate the effect of the t-scalar order and size on TPCA. We use the neighborhood strategy reported in [6, 7] to extend each image pixel to a high-order compound pixel represented by a t-scalar. Consequently, a generalized image is referred to as a compound image, which is represented by a t-matrix. Then, we adopt TPCA to analyze an obtained compound image and compare its performance, in data compression and reconstruction, with its canonical counterpart PCA.

The remainder of this paper is organized as follows. In Section \ref{section:002}, 
we introduce 
the mechanism of TPCA and its reconstruction on compound images. 
In Section \ref{section:TPCA-decomposed}, we describe of TPCA's implementation via a decomposition mechanism in the Fourier domain and two stratgies of extend a legacy monochrome image to a compound image represented by a t-matrix.  
In Section \ref{section:004}, we give the expriment results by PCA and different TPCA variants on 
public images. We conclude this paper in Section \ref{section:005}.

\section{Image Reconstruction via TPCA}
\label{section:002}

The notations, index protocols, symbols of this paper follow the existing ones of \cite{liao2020general} as much as possible. For example, all indices begin from 1 rather than 0. Different symbol subscripts other than symbol fonts are used for different data types since there are many data types rather than just canonical scalars, vectors, matrices, and tensors. Interested readers are referred to \cite{liao2020general} for more details of these symbol subscripts.  To be more precise, some example notations
with the t-matrix paradigm and their descriptions are given in table \ref{tab:table001}.

\newcommand\liaoliang[2]{
$
\begin{matrix}
\text{#1} \\
\text{#2} \\
\end{matrix}
$
}

\newcommand\liaoliangtab[2]{
\begin{tabular}{ll}
\hspace{-0.7em}\text{#1} \\
\hspace{-0.7em}\text{#2} \\
\end{tabular}
}

\begin{table}[tbh]
  \centering
  \caption{Some notations and their descriptions used in this paper}
   \vspace{-0.8em}
\resizebox{0.5\textwidth}{!}{
    \begin{tabular}{|l|l|}
    \hline 
    
    \hline
    
    \hline
    Notations & Description\\
    \hline
    \hline
    $C \equiv \mathbb{C}^{I_1\times \cdots \times I_N}$     & t-algebra of t-scalars  \\
    \hline
    $X_\mathit{vec}, X_\mathit{mat}$  & canonical vector and matrix over complex numbers\\
    \hline
    $X_\mathit{TV}, X_\mathit{TM}$    & t-vector and t-matrix, i.e., vector and matrix over $C$ \\
    \hline
    $Z_{T}, E_{T} \in C$ & zero t-scalar, identity t-scalar\\
    \hline
    $Z_\mathit{TM}, I_\mathit{TM} $ & zero t-matrix, identity t-matrix\\
    \hline
    $X_{T} + Y_{T} \in C $ & t-scalar addition\\
    \hline
    $X_\mathit{TM} + Y_\mathit{TM}$ & t-matrix addition\\
    \hline
    $X_\mathit{T} \circ Y_\mathit{T} \in C$ & t-scalar multiplication\\
    \hline
    $X_\mathit{TM} \circ Y_\mathit{TM}$ & t-matrix multiplication\\
    \hline
    $\raisebox{-0.3em}{$\tilde{X}_\mathit{TV} \doteq F(X_\mathit{TV})$}  $ & 
    \raisebox{-0.3em}{multi-way Fourier transform of a t-vector $X_\mathit{TV}$} \\
    \hline
    
    $\raisebox{-0.3em}{${X}_\mathit{TV} \doteq F^{-1}(\tilde{X}_\mathit{TV})$}  $ & 
        \raisebox{-0.3em}{inverse multi-way Fourier transform for a t-vector } \\
        \hline
    
    $\tilde{X}_\mathit{TV}(i_1,\cdots,i_N) $  & \liaoliangtab{the $(i_1,\cdots,i_N)$-th slice of a transformed }{
    t-vector in the Fourier domain}  \\
    \hline
    \raisebox{-0.5em}{\hspace{-0.5em}
    $(X_\mathit{TM})_{1:d_1,\,1:d_2}$
    }    & 
    \raisebox{-0.9em}{
    \liaoliangtab{\hspace{-0.6em}a t-matrix in $C^{d_1\times d_2}$, which perserves the first $d_1$}{\hspace{-0.6em}and the first $d_2$ columns of the t-matrix $X_\mathit{TM}$}
    }
     \\
    \hline
    $X_\mathit{TM}^{\raisebox{0em}{$*$}}$ & conjugate tranpose of a t-matrix $X_\mathit{TM}$ \\
    \hline        
    
    \hline
    
    \hline
    \end{tabular}
}
  \label{tab:table001}
\end{table}

For those notations not appearing in \cite{liao2020general}, we give their descriptions when necessary.

\subsection{PCA}

Due to image data redundancy, a raw image represented by a vector can be effectively mapped to a low-dimensional subspace. Canonical PCA seeks such low-dimensional subspace and extracts predominant image features with the help of SVD of a matrix. The specific steps can be summarized as follows.

Given $K$ training vectors $X_{\mathit{vec}, 1}, \cdots, X_{\mathit{vec}, K} \in \mathbb{R}^{D}$, their covariance matrix $G_{\mathit{mat}} \in \mathbb{R}^{D\times D}$ of these vectors is given by
\begin{equation} 
G_\mathit{mat} = \scalebox{1}{$\frac{1}{K-1}$} \cdot \scalebox{1}{$\sum\nolimits_{k=1}^{K}$} 
(X_{\mathit{vec}, k} - \bar{X}_{\mathit{vec}}) \cdot
(X_{\mathit{vec}, k} - \bar{X}_{\mathit{vec}})^{T} 
\end{equation}
where 
\begin{equation} 
\bar{X}_\mathit{vec} = (1/K) \cdot \scalebox{1}{$\sum\nolimits_{k=1}^{K}$} X_{\mathit{vec}, k}\;.
\end{equation}

It is easy to verify that $G_\mathit{mat}$ is symmetric, namely 
$G_\mathit{mat} = G_\mathit{mat}^{T}$. Let the SVD of $G_\mathit{mat}$ be 
\begin{equation}
G_\mathit{mat} = U_\mathit{mat} \cdot S_\mathit{mat} \cdot U_\mathit{mat}^{T} \,.
\end{equation}
such that the matrix $U_\mathit{mat} \in \mathbb{R}^{D\times D}$ is orthogonal; namely, 
$U_\mathit{mat}^{T} \cdot U_\mathit{mat} 
=  U_\mathit{mat} \cdot U_\mathit{mat}^{T} = I_\mathit{mat} 
$, where $I_\mathit{mat}$ denotes the identity matrix, and the matrix 
$S_\mathit{mat} \in \mathbb{R}^{D\times D}$ is diagonal.

The feature vector of a test vector $Y_\mathit{vec} \in \mathbb{R}^{D}$ can be computed via the orthogonal matrix $U_\mathit{mat}$  and the average vector $\bar{X}_\mathit{vec}$ of the training vectors, namely
\begin{equation}
Y_\mathit{vec}^\mathit{feature}=U_\mathit{mat}^T \cdot (Y_\mathit{vec} -\bar{X}_\mathit{vec} ) \in \mathbb{R}^{D}
\label{equation:PCA-feature}
\end{equation}

If one needs to reduce the dimension of the feature vector 
$Y_\mathit{vec}^\mathit{feature} \in \mathbb{R}^{D}$ from $D$ to $d$, 
the last $(D-d)$ entries of $Y_\mathit{vec}$ are just discarded.

The reconstructed vector $Y_\mathit{vec}^\mathit{recon} \in \mathbb{R}^{D}$ 
is computed from  
the feature vector $Y_\mathit{vec}^\mathit{feature}$ and the matrix $U_\mathit{mat}$ as 
follows. 
\begin{equation}
Y_\mathit{vec}^\mathit{recon} =  (U_\mathit{mat})_{:, 1:d} \cdot (Y_\mathit{vec}^\mathit{feature})_{1:d} + \bar{X}_\mathit{vec}
\label{equation:PCA-reconstruction}
\end{equation}
where $(U_\mathit{mat})_{:, 1:d}$ denotes the sub-matrix that has the first $d$ columns of 
the matrix $U_\mathit{mat}$, and $(Y_\mathit{vec}^\mathit{feature})_{1:d}$ denotes the sub-vector that 
remains the first $d$ entries of the vector $Y_\mathit{vec}^\mathit{feature}$. 

\subsection{TPCA}

TPCA is a straightforward generalization, over the t-algebra $C$, of the canonical PCA 
\cite{liao2020generalized, liao2020general} and summarized as follows.

Given the training t-vectors $X_{\mathit{TV},\,1},\cdots,X_{\mathit{TV},\,K} \in C^{D} \equiv \mathbb{C}^{I_1\times \cdots \times I_N\times D}$, the covariance t-matrix
$G_{TM} \in C^{D\times D}\equiv \mathbb{C}^{I_1\times \cdots \times I_N \times D\times D} $ 
is given by 
\begin{equation}
\resizebox{0.44\textwidth}{!}{$
\left\{
\begin{aligned}
&G_\mathit{TM} = \frac{1}{K-1} \cdot \scalebox{1}{$\sum\nolimits_{k=1}^{K}$}
(X_{\mathit{TV},\,k} - \bar{X}_\mathit{TV}) \circ (X_{\mathit{TV},\,k} - \bar{X}_\mathit{TV})^{\raisebox{0em}{$*$}}
 \\
&\bar{X}_\mathit{TV} = ({1}/{K}) \cdot \scalebox{1}{$\sum\nolimits_{k=1}^{K}$} 
X_{\mathit{TV},\,k}
\end{aligned}
\right.
$}
\end{equation}
where $\circ$ denotes the t-matrix multiplication defined via the multi-way circular convolution at the t-scalar entry level. Interested readers are referred to the following text for the relevant definitions \cite{liao2020generalized,liao2020general}.

Let $G_\mathit{TM} = U_\mathit{TM} \circ S_\mathit{TM} \circ U_\mathit{TM}^{\raisebox{0em}{$*$}}$
be 
the TSVD of the t-matrix $G_\mathit{TM} \in C^{D\times D}\equiv \mathbb{C}^{I_1\times \cdots I_N \times D\times D}$ such that 
\begin{equation}
\begin{aligned}
&U_\mathit{TM}^{\raisebox{0em}{$*$}} \circ U_\mathit{TM} = I_\mathit{TM}
\equiv \operatorname{diag}(E_{T},\cdots,E_{T})   \\ 
&\hspace{5.2em}\in C^{D\times D}\equiv \mathbb{C}^{I_1\times \cdots \times I_N \times D\times D} \,,\\
&S_\mathit{TM} \doteq \operatorname{diag}(\lambda_{T,\,1},\cdots,\lambda_{T,\,D}) \\ 
&\hspace{2em} \in C^{D\times D} \equiv \mathbb{C}^{I_1\times \cdots I_N\times D\times D} \,.
\end{aligned} 
\end{equation}

Then, the orthogonal t-matrix $U_\mathit{TM} \in C^{D\times D}$ and the average t-vector $\bar{X}_\mathit{TV} \in C^{D}$ can be used to reconstruct the t-vector $Y_\mathit{TV}^\mathit{recon} \in C\equiv \mathbb{C}^{I_1\times \cdots\times I_N\times D}$, an approximation to the input t-vector $Y_\mathit{TV} \in C \equiv \mathbb{C}^{I_1\times \cdots \times I_N\times D}$. More precisely, one thas the following equations. 
\begin{equation}
Y_\mathit{TV}^\mathit{feature} = U_\mathit{TM}^{\raisebox{0em}{$*$}} \circ (Y_\mathit{TV} -\bar{X}_\mathit{TV} )
\label{equation:TPCA-features}
\end{equation}
and 
\begin{equation}
Y_\mathit{TV}^\mathit{recon} = (U_\mathit{TM})_{:, 1:d}  \circ (Y_\mathit{TV}^\mathit{feature}  )_{1:d} + \bar{X}_\mathit{TV}
\label{equation:TPCA-reconstruction}
\end{equation}
where $(U_\mathit{TM})_{:,\,1:d} \in C^{D\times d} \equiv \mathbb{C}^{I_1\times \cdots\times I_N \times D\times d}$ denotes the sub-t-matrix which remains the first $d$ columns (i.e., t-vectors) of the t-matrix $U_\mathit{TM} \in C^{D\times D}$, and $(Y_\mathit{TV}^\mathit{feature} )_{1:d} \in C^{d}$ denotes the sub-t-vector which remains
the first $d$ t-scalar entries of $Y_\mathit{TV}^\mathit{feature}$.  

Using the t-matrix paradigm reported in \cite{liao2020general}, one can verify that when $I_1 = \cdots = I_N = 1$, all t-vectors and t-matrices reduce to their canonical versions, i.e., canonical vectors and matrices. 
Furthermore, 
equations (\ref{equation:TPCA-features})
and (\ref{equation:TPCA-reconstruction}) respectively reduce to equations (\ref{equation:PCA-feature}) and 
(\ref{equation:PCA-reconstruction}). Namely, in this scenarion, TPCA reduces to PCA.

\section{TPCA Decomposed and Beyond}
\label{section:TPCA-decomposed}
We describe in this section the mechanism to t-matricize a legacy image to its high-order counterpart, i.e., a compound image, and decompose TPCA to a finite number of canonical constituents.

\subsection{Compound pixels}

With the semisimple paradigm over the t-algebra in \cite{liao2020generalized,liao2020general}, t-scalars are represented by fixed-sized order-$N$ arrays in $C \equiv \mathbb{C}^{I_1\times \cdots \times I_N}$.

A $D_1\times D_2$ matrix over $C$, i.e., a $D_1\times D_2$ matrix with t-scalar
entries, is represented by an order-$(N+2)$ array in $C^{D_1\times D_2} 
\equiv \mathbb{C}^{I_1\times \cdots \times I_N \times D_1\times D_2}$. 
A matrix over $C$ is called a ``t-matrix''.  
A vector over $C$ is called a ``t-vector'', which is a special case of t-matrix. 

T-matrices can be scaled, added, scaled, multiplied, conjugate transposed, inverted or pseudo-inverted,  in a way backward-compatible  to their canonical counterparts defined over complex numbers \cite{liao2020general}. 

To use the t-matrix paradigm for general image analysis, one needs a mechanism to 
extend a pixel, represented by a scalar, to its high-order version, i.e., a compound pixel represented by a t-scalar in the form of an order-$N$ array. 

Such an extension from one scalar (pixel) to a t-scalar (compound pixel) can be realized using a fixed-sized neighborhood of each pixel.  An $D_1\times D_2$ image with compound pixel is called a ``compound image'' and represented by a t-matrix in $C^{D_1\times D_2} \equiv \mathbb{C}^{I_1\times \cdots \times I_N \times D_1\times D_2}$.

We use two pixel neighborhood strategies to extend each pixel to its high-order version, i.e., a compound pixel.

\begin{figure}[tbh]
\centering
\includegraphics[width = 0.5\textwidth]{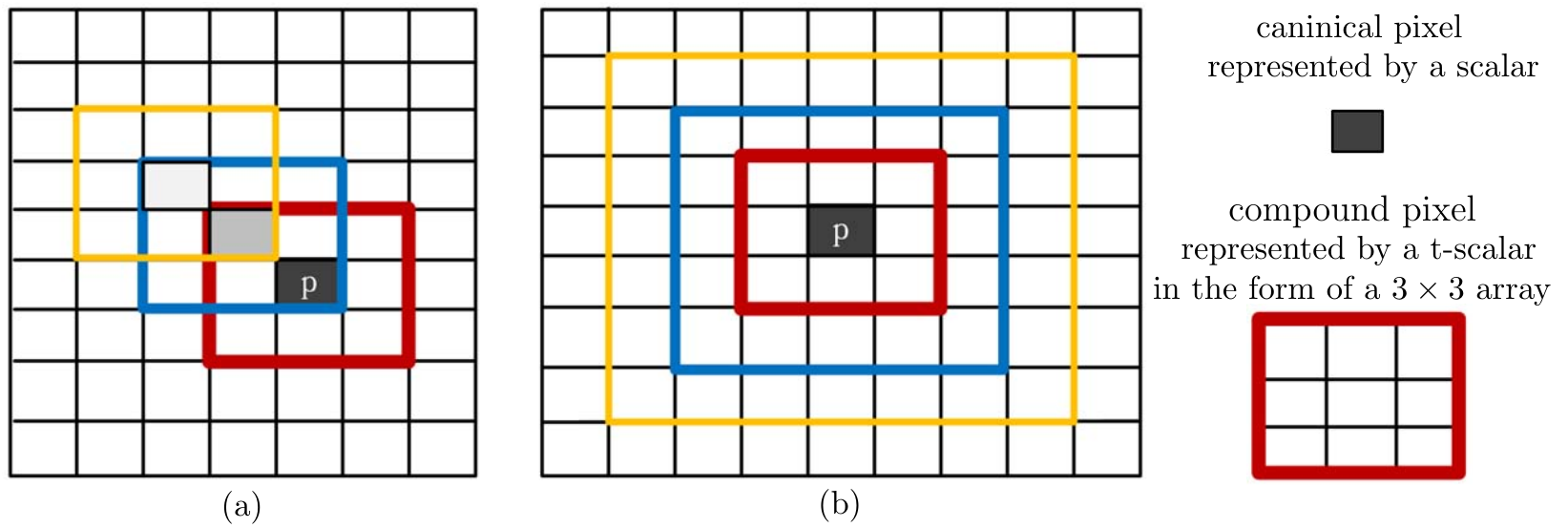}
\caption{Two strategies of using pixel neighborhoods to extend each pixel to a compound pixel}
\label{fig:neighborhood-strategy}
\end{figure}

\textbf{Compound pixel strategy 1}.\;
The first strategy is shown in figure \ref{fig:neighborhood-strategy}(a). Given a canonical pixel $p$, one can find a $3\times 3$ neighborhood centered at $p$. 
A red-line box in the subfigure highlights the $3\times 3$ neighborhood.  

Now, one extends a scalar/pixel to an order-two t-scalar, i.e., a $3\times 3$ array.

One can reuse the $3\times 3$ neighborhood  strategy 
to increase the order of t-scalars further.Each pixel (scalar) in the red-line box has a $3\times 3$ neighborhood. For example, the top-left pixel (scalar) in the red-line box has a $3\times 3$ neighborhood highlighted by a blue-line box.   

When one can substitutes each pixel (scalar) in the red box, one can have an order-four 
array of size $3\times 3\times  3\times 3$. 
Similarly, after a third reusing the $3\times 3$ neighborhood strategy, one can extend a pixel  (scalar) $p$ to its order-six version, i.e., a compound pixel (t-scalar) in the form of a $3\times 3\times 3\times 3\times 3\times 3$ array.   

\textbf{Compound pixel strategy 2}.\;
The second strategy is shown in figure \ref{fig:neighborhood-strategy}(b). This strategy 
increases the size of the array rather than its order. The t-scalar extension for the pixel $p$ is always in the form of an order-two array. However, one can use a $3\times3$ neighborhood, $5\times 5$ neighborhood, $7\times 7$ neighborhood
or $9\times 9$
neighborhood, etc., for the pixel $p$. 

If a scalar (pixel) is located at the image border, one can pad with $0$ when
necessary to have a neighborhood.  
No matter using which strategy, when the size of t-scalars is determined, it is fixed and should not be changed anymore. 

Analogous to the process of recasting an image to a vector, one can recast an obtained compound image to a t-vector.

Using the multi-way Fourier transform described in \cite{liao2020generalized} and \cite{liao2020general}, one can decompose any t-vector/t-matrix operation to 
a finite number of canonical vector/matrix counterparts carried out in the Fourier domain. 

TPCA can be efficiently implemented in this way using the multi-way Fourier transform on input data. More precisely, TPCA is given by Algorithm \ref{algorithm:TPCA}. 
 
\textbf{Fourier slice}\;.
In Algorithm \ref{algorithm:TPCA} we use the so-called Fourier slice mechanism to decompose a fourier-transformed t-matrix $\tilde{X}_\mathit{TM} \in C^{D_1\times D_2} \equiv 
\mathbb{C}^{I_1\times \cdots \times I_N \times D_1\times D_2}$ to $K^\mathit{slice}$ complex matrices  
$
\tilde{X}_\mathit{TM}(i_1, \cdots, i_N) \in \mathbb{C}^{D_1\times D_2}
$ for all $(i_1, \cdots, i_N) \in [I_1]\times \cdots \times [I_N]$,
where $K^\mathit{slice} \doteq I_1\cdots I_N$ such that the following condition holds
for all $(d_1, d_2) \in [D_1]\times [D_2]$, 
\begin{equation} 
\resizebox{.43\textwidth}{!}{$
\scalebox{1.4}{$($}
\tilde{X}_\mathit{TM}(i_1,\cdots,i_N)
\scalebox{1.4}{$)$}_{d_1, d_2} = 
\scalebox{1.4}{$($}
\scalebox{1}{$($}
\tilde{X}_\mathit{TM}
\scalebox{1}{$)$}_{d_1, d_2}  
\scalebox{1.4}{$)$}_{i_1,\cdots,i_N} \in \mathbb{C}
\;.
$}
\end{equation}
where $(\tilde{X}_\mathit{TM})_{d_1,d_2} \in C \equiv \mathbb{C}^{I_1\times \cdots \times I_N}$.

When $D_2 = 1$, the fourier-transformed t-matrix $\tilde{X}_\mathit{TM}$ reduces to a fourier-transformed t-vector $\tilde{X}_\mathit{TV}$. Therefore, one has the fourier
slice $\tilde{X}_\mathit{TV}(i_1,\cdots,i_N) $ for all $(i_1,\cdots,i_N) \in [I_1]\times \cdots \times [I_N]$.

\renewcommand{\algorithmicrequire}{\textbf{Input:}}
\renewcommand\algorithmicensure {\textbf{Output:} }
\begin{algorithm}[tbh]
\small
\caption{Principal features extraction and data reconstruction via TPCA for t-vectors}
\begin{algorithmic}[1]
\REQUIRE training t-vectors $X_{\mathit{TV},\,1},\cdots,X_{\mathit{TV},\,K}$ and a query t-vector  $Y_\mathit{TV}$ in $C^{D}$ $\equiv$ $\mathbb{C}^{I_1\times \cdots \times I_N \times D}$
\ENSURE 
feature t-vector $Y_\mathit{TV}^\mathit{feature} \in C^{D} \equiv \mathbb{C}^{I_1\times \cdots \times I_N \times D}$ as in equation (\ref{equation:TPCA-features}) and 
reconstructed t-vector $Y_\mathit{TV}^\mathit{recon}$ $\in$ $C^{D}$ $\equiv$ 
$C^{I_1\times \cdots \times I_N\times D} $ as in equation (\ref{equation:TPCA-reconstruction}).
\STATE  Compute the following Fourier transforms 
\[
\tilde{X}_{\mathit{TV},\,k} \leftarrow F(X_{\mathit{TV}\,k}),\;\forall k \in [K]
\;,\;\;
\tilde{Y}_{\mathit{TV}} \leftarrow F(Y_{\mathit{TV}})\;.
\]
\FORALL {$(i_1,\cdots,i_N) \in [I_1]\times \dots \times [I_N]$  }
\STATE Compute the following covariance matrix $G_\mathit{mat} \in \mathbb{C}^{D\times D}$. 
\[ 
\resizebox{0.46\textwidth}{!}{$
G_\mathit{mat} = \scalebox{1}{$\frac{1}{K-1}$} \cdot \scalebox{1}{$\sum\nolimits_{k=1}^{K}$}  
\big(\scalebox{0.9}{$\tilde{X}_\mathit{TV}(i_1,\cdots,i_N) - \bar{X}_\mathit{vec} $}  \big) \cdot 
\big(\scalebox{0.9}{$\tilde{X}_\mathit{TV}(i_1,\cdots,i_N) - \bar{X}_\mathit{vec}$}  \big)^{H}
$}
\]
where 
$
\bar{X}_\mathit{vec} = \scalebox{1}{$\frac{1}{K}\cdot \sum\nolimits_{k=1}^{K}$} 
\tilde{X}_\mathit{TV}(i_1,\cdots,i_N) \in \mathbb{C}^{D} \;.
$
\STATE  Compute the canonical SVD of the hermitian  matrix $G_\mathit{mat} \in \mathbb{C}^{D\times D}$ such that 
\[
G_\mathit{mat} = U_\mathit{mat} \cdot S_\mathit{mat} \cdot U_\mathit{mat}^{H} 
\]
where $U_\mathit{mat} \in \mathbb{C}^{D\times D}$, $S_\mathit{mat} \in \mathbb{C}^{D\times D}$.
\STATE  Compute the $(i_1,\cdots,i_N)$-th fourier slice of the fourier-transformed t-vectors $\tilde{Y}_\mathit{TV}^\mathit{feature}, \tilde{Y}_\mathit{TV}^\mathit{recon} \in \mathbb{C}^{I_1\times \cdots I_N\times D}$, more precisely, 
\[
\tilde{Y}_\mathit{TV}^\mathit{feature}(i_1,\cdots,i_N) \leftarrow U_\mathit{mat}^{H}
\cdot 
\scalebox{1.3}{$($}
\tilde{Y}_\mathit{TV}(i_1,\cdots,i_N) - \bar{X}_\mathit{vec}
\scalebox{1.3}{$)$}
\]
\[
\resizebox{0.45\textwidth}{!}{$
\tilde{Y}_\mathit{TV}^\mathit{recon}(i_1,\cdots,i_N) \leftarrow (U_\mathit{mat})_{:, 1:d} \cdot 
\scalebox{1.3}{$($}
\tilde{Y}_\mathit{TV}^\mathit{feature}(i_1,\cdots,i_N) 
\scalebox{1.3}{$)$}_{1:d} + \bar{X}_\mathit{vec}
$}
\]
\ENDFOR
\STATE \textbf{return} $Y_\mathit{TV}^\mathit{feature} \in C^{D} \equiv \mathbb{C}^{I_1\times \cdots \times I_N\times D}$
and $Y_\mathit{TV}^\mathit{recon} \in C^{D} \equiv \mathbb{C}^{I_1\times \cdots \times I_N \times D} $ give by 
$Y_\mathit{TV}^\mathit{feature} \leftarrow F^{-1}(
\scalebox{0.9}{$\tilde{Y}_\mathit{TV}^\mathit{feature}$} 
)$ and 
$
Y_\mathit{TV}^\mathit{recon} \leftarrow F^{-1}(
\scalebox{0.9}{$\tilde{Y}_\mathit{TV}^\mathit{recon}$}) \;.
$
\end{algorithmic}
\label{algorithm:TPCA}
\end{algorithm}

\section{Experimental Results and Analysis}
\label{section:004}

We use PCA and TPCA on the public MNIST handwritten digit dataset for image reconstruction experiments. Each image of the MNIST  dataset is a monochrome image of the size $28\times 28$. There are ten classes in the dataset, i.e., the classes respectively for the digits $0,\cdots,9$.  We randomly choose $60$ images from each class as training images and $10$ images from each class as quey images.

We compare the reconstruction performances of PCA and TPCA with different t-scalar shapes.

Using or reusing the neighborhood of each pixel described as in strategies 1 and 2, one can have different t-scalar shapes. In this section, TPCA with different t-scalar shapes is respectively named as TPCA, TPCA-A, TPCA-B (TPCA variants using compound pixel strategy  1) and TPCA-X, TPCA-Y, TPCA-Z (TPCA variants using compound pixel strategy 2). 

When each t-scalar only contains one scalar entry (in other words, t-scalars are order-zero), TPCA reduces to PCA. Namely, PCA is a special TPCA with order-zero t-scalars.   
The detailed t-scalar settings are given in 
table \ref{tab:TPCA-with-different-tsize}.

\begin{table}[t]
\caption{TPCA with different shapes of t-scalars}
    \vspace{-0.8em}
\begin{minipage}{1\textwidth}
\resizebox{0.5\textwidth}{!}{
    \begin{tabular}{|c|c|c|c|}
    \hline 
    
    \hline
    
    \hline
    \liaoliang{\raisebox{-0.1em}{TPCA}}{\raisebox{0.1em}{variant}}  & \liaoliang{\raisebox{-0.1em}{t-scalar shape}}{\raisebox{0.1em}{(i.e., $I_1\times \cdots\times I_N$)}}  & \liaoliang{\raisebox{-0.1em}{t-scalar}}{\raisebox{0.1em}{order}} & \multicolumn{1}{l|}{~~~\liaoliang{\raisebox{-0.1em}{neighborhood}}{\raisebox{0.1em}{strategy}} } \\
    \hline    
    \hline
    \raisebox{-0.1em}{PCA} & \raisebox{-0.1em}{$1$} & \raisebox{-0.1em}{$0$} & \raisebox{-0.1em}{N/A}\\
    \hline
    \hline
    \raisebox{-0.1em}{TPCA}     & \raisebox{-0.1em}{$3\times3$}     & \raisebox{-0.1em}{$2$}     & \multicolumn{1}{c|}{\multirow{3}[0]{*}{\liaoliang{compound pixel}{strategy 1}}} \\
    \cline{1-3}
    \raisebox{-0.1em}{TPCA-A}     & \raisebox{-0.1em}{$3\times3 \times 3\times 3$}     & \raisebox{-0.1em}{$4$}     & \multicolumn{1}{c|}{} \\
    \cline{1-3}
    \raisebox{-0.1em}{TPCA-B}     & \raisebox{-0.1em}{$3\times3 \times 3\times 3\times 3\times 3$}      & \raisebox{-0.1em}{$6$}     & \multicolumn{1}{c|}{} \\
    \hline
    \hline
    \raisebox{-0.1em}{TPCA-X}     & \raisebox{-0.1em}{$5\times 5$}     & \raisebox{-0.1em}{$2$}     & \multicolumn{1}{c|}{\multirow{3}[0]{*}{\liaoliang{compound pixel}{strategy 2}}} \\
    \cline{1-3}
    \raisebox{-0.1em}{TPCA-Y}     & \raisebox{-0.1em}{$7\times 7$}     & \raisebox{-0.1em}{$2$}     & \multicolumn{1}{c|}{} \\
    \cline{1-3}
    \raisebox{-0.1em}{TPCA-Z}     & \raisebox{-0.1em}{$9\times 9$}     & \raisebox{-0.1em}{$2$}     & \multicolumn{1}{c|}{} \\
    \hline
    
    \hline
    
    \hline
    \end{tabular}%
}
  \label{tab:TPCA-with-different-tsize}%
\end{minipage}
\end{table}

The reconstruxtion quality is measured in PSNR (Peak Signal Noise Ratio). More precisely, given an array $X$ and its approximation version $X^\mathit{recon}$, their PSNR is given as follows. 
\begin{equation} 
\mathit{PSNR} = 20 \log_{10} \frac{\mathit{MAX} \cdot \sqrt{D} }
{\|X - X^\mathit{recon}\|_F}
\label{equation:PSNR}
\end{equation}
where $\mathit{MAX}$ denotes the largest possible value of the entries in $X$ and $D$ denotes the number of entry of $X$, and $\|\cdot \|_F$ stands for the 
Frobenius norm of an array.   
In this paper, the parameter 
$\mathit{MAX}$ is a constant, i.e., $\mathit{MAX} = 255$. 

Each raw MNIST image is a $28\times 28$ array, i.e., with $784$ pixels.  
When using PCA, each raw or 
reconstructed MNIST image is recast to a vector in $\mathbb{R}^{784}$.  
Hence, one can organize $100$ raw or reconstructed vectors, as columns, in an array in $\mathbb{R}^{784\times 100}$  
As a consequence, $100$ raw images form an array $X \in \mathbb{R}^{784\times 100}$ and 100 recontructed vectors form an array $X^\mathit{recon}$ of the same size. 

By substituting the arrays $X$ and $X^\mathit{recon}$ 
to equation (\ref{equation:PSNR}) 
with $D = 78400 \equiv 784\times 100 $, one can have the PSNR of $X$ annd $X^\mathit{recon}$.

Note that TPCA with different shapes of t-scalars yields t-vectors rather than vectors. 
Each t-vector represents a compound image and is in the form of a high-order array in $C^{D} \equiv  \mathbb{C}^{I_1\times \cdots \times I_N\times D}$. Our experiment setting is given by table \ref{tab:TPCA-with-different-tsize} with 
the shape of t-scalars being 
$I_1 \times \cdots \times I_N$.

To give a fair comparison of the performances of PCA and TPCA, we extract the cental spatial slices of the input and output t-vectors yielded by TPCA for computing PSNRs.  

A spatial slice is analogous to a Fourier slice described in Section \ref{section:TPCA-decomposed} but extracted in the spatial domain rather than the Fourier domain. 

The $(i_1,\cdots,i_N)$-th spatial slice of a t-vector  $X_\mathit{TV} \in C^{D} \equiv \mathbb{C}^{I_1\times \cdots \times I_N\times D}$ is denoted by 
$
X_\mathit{TV}(i_1,\cdots,i_N) \in \mathbb{C}^{D}
$ such that, for all 
$(i_,\cdots,i_N, d) \in [I_1]\times \cdots \times [I_N] \times [D]$, the following equality holds  
\begin{equation}
\scalebox{1.3}{$($}  X_\mathit{TV}(i_1,\cdots,i_N)\scalebox{1.3}{$)$} _{d} = 
\scalebox{1.3}{$($}
\scalebox{1}{$($}
X_\mathit{TV}
\scalebox{1}{$)$}_{d} 
\scalebox{1.3}{$)$}_{i_1,\cdots,i_N} \in \mathbb{C}
\;\;. 
\end{equation}
where $
\scalebox{1}{$($}X_\mathit{TV}
\scalebox{1}{$)$}_{d} \in C \equiv \mathbb{C}^{I_1\times \cdots \times I_N}$.

In our experiments, the shape of t-scalars 
is given by $I_1 \times \cdots \times I_N$ where $I_1,\cdots,I_N$ are odd positive integers. Therefore, the so-called 
central spatial slice of a t-vector $X_\mathit{TV} \in C^{D} \equiv \mathbb{C}^{I_1\times \cdots\times I_N \times D}$ is given by $
X_\mathit{TV}(\frac{I_1+1}{2},\cdots,\frac{I_N+1}{2}) \in \mathbb{C}^{D} $ with $D = 784$ in our experiments. 

The input t-vector slices and output reconstructed t-vector slices are respectively organized in two $784\times 100$  arrays $X$ and $X^\mathit{recon}$. The two arrays are substituted in 
equation (\ref{equation:PSNR}) with $D = 78400$ for computing PSNRs.

\begin{figure}[h]
\centering
    \includegraphics[width=0.46\textwidth]{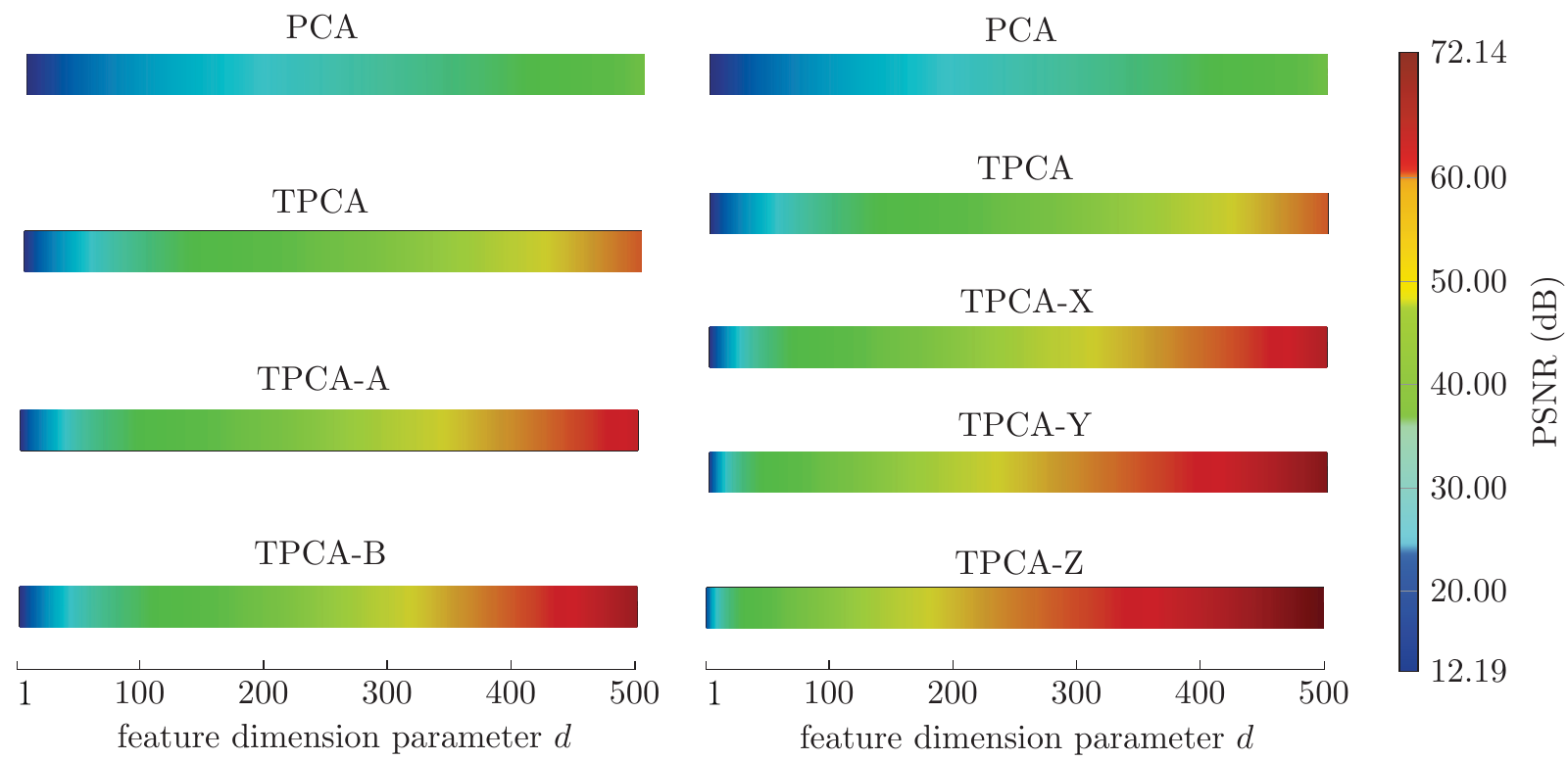}\\
    \caption{A heat-map comparison of PSNRs by PCA and different TPCA variants}
\label{fig:visual-comparison-PCA-TPCA}
\end{figure}

Some quantitative PSNR comparsion of PCA and different TPCA variants is given by table \ref{tab:PSRN-comparison}. 
A visual comparison of the PSNRs by PCA and different TPCA variants is given by figure \ref{fig:visual-comparison-PCA-TPCA}.

The PSNRs in table \ref{tab:PSRN-comparison} and figure \ref{fig:visual-comparison-PCA-TPCA}
are outputs of PCA and TPCA with the feature dimension parameter $d = 50, 100, \cdots, 500$.

\newcommand\liaoliangthree[3]{
$
\begin{matrix}
\text{#1} \\
\text{#2} \\
\text{#3} 
\end{matrix}
$
}

\begin{table*}[htb]
  \centering
  \caption{PSNR comparison of PCA and different TPCA variants}
\vspace{-0.8em}
\resizebox{1\textwidth}{!}{
    \begin{tabular}{|c||c|c||c|c||c|c|c|}
    \hline
    
    \hline
    
    \hline 
    \liaoliangthree{feature}{dimension}{parameter $d$} & \scalebox{0.9}{PCA}    & \scalebox{0.9}{\liaoliang{TPCA}{($3\times 3$ t-scalars)} }  & 
    \scalebox{0.9}{\liaoliangthree{TPCA-A}{($3\times 3\times  3\times 3$}{t-scalars)} } & 
    \scalebox{0.9}{\liaoliangthree{TPCA-B}{($3\times 3\times  3\times 3\times  3\times 3$}{ t-scalars)} }
     & \scalebox{0.9}{\liaoliang{TPCA-X}{($5\times 5$ t-scalars)}} 
     & \scalebox{0.9}{\liaoliang{TPCA-Y}{($7\times 7$ t-scalars)}} 
     & \scalebox{0.9}{\liaoliang{TPCA-Z}{($9\times 9$ t-scalars)}} \\
    \hline\hline 
    $50$    & $\numprint{18.6101}$ & $\numprint{23.4491}$ & $\numprint{26.304}$ & $\numprint{25.2405}$ & $\numprint{29.0696}$ & $\numprint{32.5606}$ & $\numprint{35.0611}$ \\
    \hline 
    $100$   & $\numprint{21.2687}$ & $\numprint{28.646}$ & $\numprint{31.8457}$ & $\numprint{30.6851}$ & $\numprint{34.4083}$ & $\numprint{38.1932}$ & $\numprint{41.1955}$ \\
    \hline 
    $150$   & $\numprint{22.9703}$ & $\numprint{32.4564}$ & $\numprint{35.6941}$ & $\numprint{34.8035}$ & $\numprint{38.2627}$ & $\numprint{42.255}$ & $\numprint{45.6282}$ \\
    \hline 
    $200$   & $\numprint{24.3531}$ & $\numprint{35.3664}$ & $\numprint{38.9333}$ & $\numprint{38.8171}$ & $\numprint{41.5191}$ & $\numprint{45.8779}$ & $\numprint{49.5811}$ \\
    \hline 
    $250$   & $\numprint{25.7561}$ & $\numprint{38.1016}$ & $\numprint{42.192}$ & $\numprint{42.7546}$ & $\numprint{44.6509}$ & $\numprint{49.229}$ & $\numprint{52.9634}$ \\
    \hline 
    $300$   & $\numprint{27.2613}$ & $\numprint{40.8319}$ & $\numprint{45.441}$ & $\numprint{46.7613}$ & $\numprint{47.5934}$ & $\numprint{52.2855}$ & $\numprint{55.9912}$ \\
    \hline 
    $350$   & $\numprint{29.077}$ & $\numprint{43.5783}$ & $\numprint{48.7564}$ & $\numprint{51.0266}$ & $\numprint{50.5876}$ & $\numprint{55.5792}$ & $\numprint{59.1653}$ \\
    \hline 
    $400$   & $\numprint{31.2232}$ & $\numprint{46.7117}$ & $\numprint{52.1646}$ & $\numprint{55.2362}$ & $\numprint{54.0705}$ & $\numprint{58.8976}$ & $\numprint{62.5735}$ \\
    \hline 
    $450$   & $\numprint{34.1248}$ & $\numprint{50.4626}$ & $\numprint{56.2703}$ & $\numprint{60.0806}$ & $\numprint{58.0445}$ & $\numprint{63.078}$ & $\numprint{66.7265}$ \\
    \hline
    $500$   & $\numprint{37.967}$ & $\numprint{55.2625}$ & $\numprint{61.2684}$ & $\numprint{66.0674}$ & $\numprint{63.7257}$ & $\numprint{68.6433}$ & $\numprint{72.1409}$ \\
    \hline
    
    \hline
    
    \hline
    \end{tabular}%
  \label{tab:PSRN-comparison}
} 
\end{table*}

It shows from 
table \ref{tab:PSRN-comparison} and 
figure \ref{fig:visual-comparison-PCA-TPCA}  that when the feature dimension parameter $d$ increases, the reconstruction qualities in terms of PSNRs of PCA and different TPCA variants all increase. 

Another observation from table \ref{tab:PSRN-comparison} and figure \ref{fig:visual-comparison-PCA-TPCA} is that TPCA always outperforms its canonical counterpart PCA. When $d = 500$, TPCA outperforms PCA for at least $17.296$ dB (i.e., $55.263$ dB- $37.967$ dB).

A third observation from the table and figure is that the size of t-scalars effect TPCA’s performance, sometimes significantly. For example, a complicated t-scalar shape, such as a $3\times 3\times 3 \times 3$ t-scalar shape adopted by TPCA-A, affects the performance favorably than the simple $3\times 3$ t-scalar shape adopted by TPCA.

The results in table \ref{tab:PSRN-comparison} and figure \ref{fig:visual-comparison-PCA-TPCA} also show that, in the experiment of compound image approximations, TPCA using larger pixel neighborhoods might be more effective than TPCA using higher-order pixel neighborhoods. 

For example, TPCA-A and TPCA-Z have the same amount of fourier slices since $3\times 3\times 3\times 3 = 9\times 9$. However, with the same parameter $d$, TPCA-Z yields more favorable results than TPCA-A.

\begin{figure*}[htb]
\begin{minipage}[bt] {\textwidth}
    \includegraphics[width=0.98\textwidth]{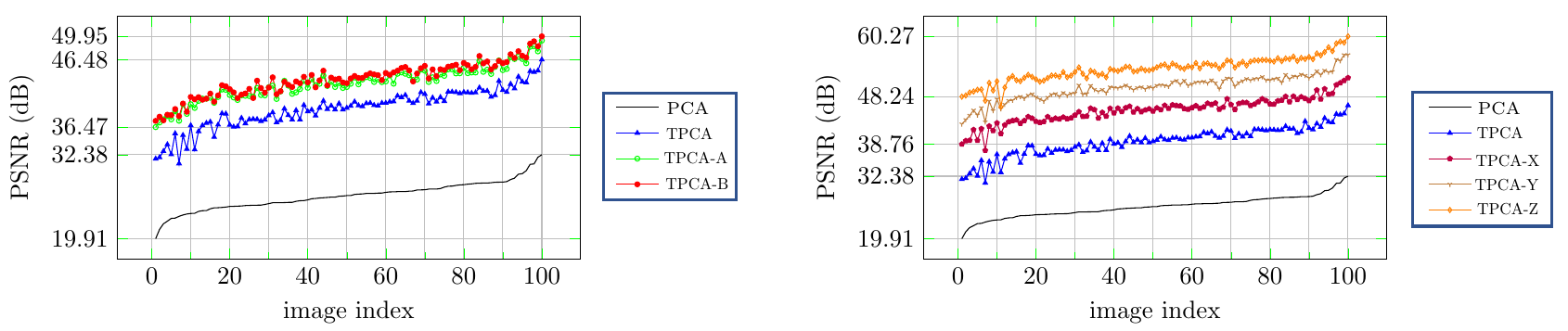}\\
    \vspace{-1.5em}
    \caption{PSNR comparison on each image reconstructed by PCA and different TPCA variants, feature dimension $d = 250$  }
    \label{figure:reconstruction-curve-on-each-imge-d250}
\end{minipage}

\vspace{1em}
\begin{minipage}[h] {\textwidth}
    \includegraphics[width=0.98\textwidth]{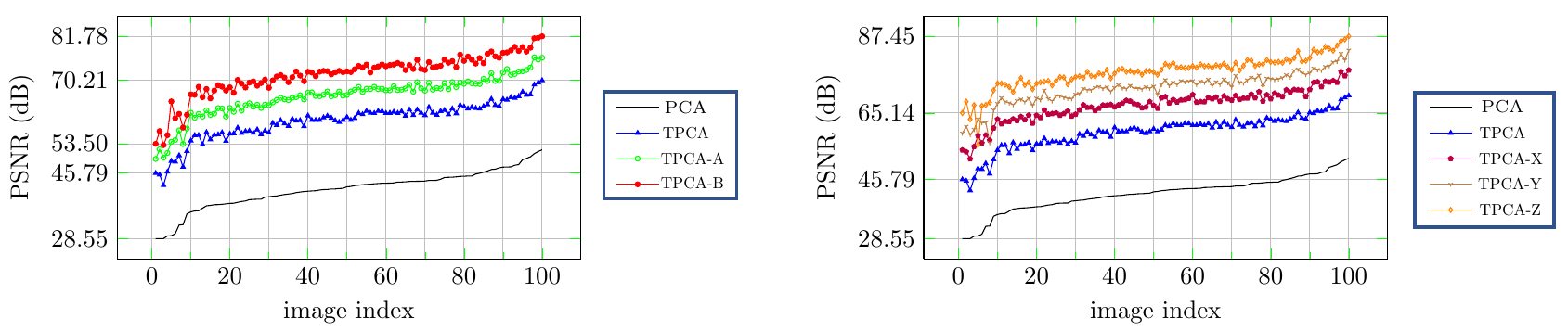}\\
    \vspace{-1.5em}
    \caption{PSNR comparison on each image reconstructed by PCA and different TPCA variants, feature dimension $d = 500$}
\label{figure:reconstruction-curve-on-each-imge-d500}
\end{minipage}
\end{figure*}

One might be interested in the reconstruction performance of PCA and TPCA on each image. 
Figures \ref{figure:reconstruction-curve-on-each-imge-d250} and \ref{figure:reconstruction-curve-on-each-imge-d500} show the PSNR curves of PCA and TPCA over each image.

The variables of the horizontal axes of figures \ref{figure:reconstruction-curve-on-each-imge-d250} and 
\ref{figure:reconstruction-curve-on-each-imge-d500} are image indices, which are sorted to make a PSNR curve of PCA an increasing function of the image index.

Figure \ref{figure:reconstruction-curve-on-each-imge-d250} corresponds to the setting of the feature dimension $d = 250$. 
Figure \ref{figure:reconstruction-curve-on-each-imge-d500} corresponds to the setting of the feature dimension $d = 500$.  
Both figures corroborate the observations found in table \ref{tab:PSRN-comparison} and figure \ref{fig:visual-comparison-PCA-TPCA}. Namely, TPCA outperforms PCA in terms of reconstruction quality.

\section{Conclusion}
\label{section:005}

The performance of generalized image reconstruction, using TPCA (Tensorial Principal Component Analysis), over a recently-reported commutative algebra, called t-algebra (tensorial algebra) \cite{liao2020generalized,liao2020general} is evaluated.  

The t-algebra generalizes the field of complex numbers and contains the generalized complex numbers referred to as t-scalars. Over the t-algebra, one can have a novel matrix paradigm that is backward-compatible to the canonical matrix paradigm. This paradigm shift might bring new insights into general data analytics, particularly into visual information analysis, including but not limited to image reconstruction. 

To validate this claim, we conduct some numerical experiments on public images using TPCA for image reconstruction. The classical PCA is a special case of TPCA. The experiments show that TPCA with different shapes of t-scalars consistently outperforms its canonical counterpart PCA. This paper also reports the effect of the shape of t-scalars on the performance of TPCA. 

For a more rigorous introduction of the t-algebra and the matrix paradigm over the algebra, interested readers are referred to \cite{liao2020generalized,liao2020general} for more details.

\section*{Acknowledgment}
This work was partially supported by the National Natural Science Foundation of China under the grant number No. U1404607, the High-end Foreign Experts Program, under the grant numbers No. GDW20186300351 and G20200226015, of the State Administration of Foreign Experts Affairs, and the Open 
Research Program of the National Engineering Laboratory for Integrated Aero-Space-Ground-Ocean Big Data Application Technology under the grant number 
No. 20200206.

Liang Liao and Stephen John Maybank contribute equally to the theory of t-algebra, t-scalars, t-vectors, and t-matrices. Liang Liao would like to thank the Birkbeck Institute of Data Analytics for free using the high-performance computing facilities during the stay in London, although London’s lockdown somehow affects this research's progress. 

This work serves as a numerical evaluation of the theory of t-algebra and the data analytics paradigm over t-algebra. Xuechun Zhang is reponsible for all experiment results. 

\section*{Open Source}
A MATLAB repository on the t-algebra, t-vectors, and t-matrices is open-sourced at 
the following URL.

www. github/liaoliang2020/talgebra. 

Interested readers are referred to it for more details. 

\balance
\bibliographystyle{IEEEtran}
\bibliography{liaoliang}

\end{document}